%% file: main.tex
\documentclass[runningheads]{llncs}

\usepackage{eccv}
\input{preamble}

\usepackage{eccvabbrv}

\usepackage{graphicx}
\usepackage{booktabs}

\usepackage[accsupp]{axessibility}  

\usepackage{hyperref}
\newcommand\rurl[1]{%
\href{https://#1}{\nolinkurl{#1}}%
}
\usepackage{orcidlink}

\begin{document}

\title{Interaction-Aware 4D Gaussian Splatting for Dynamic Hand-Object Interaction Reconstruction} 

\titlerunning{Interaction-Aware 4D-GS for Dynamic HOI Reconstruction}

\author{Hao Tian\inst{1,2,5}\orcidlink{0009-0001-8304-8549} \and
Chenyangguang Zhang\inst{3} \and
Rui Liu\inst{4} \and \\
Wen Shen\inst{1,5}\and
Xiaolin Qin\inst{1,5}\thanks{Corresponding author.}}

\authorrunning{H.~Tian et al.}


\institute{Chengdu Institute of Computer Applications, Chinese Academy of Sciences, Chengdu, China \and PetroChina (Beijing) Digital Intelligence Research Institute Co., Ltd., Beijing, China \and
Tsinghua University, Beijing, China \and 
Minzu University of China, Beijing, China \and
University of Chinese Academy of Sciences, Beijing, China \\
\rurl{haotian23ucas.github.io/Interaction-Aware-Gau}}

\maketitle

\input{LaTeX/sections/0_abstract}
\input{LaTeX/sections/1_introduction}
\input{LaTeX/sections/2_related_work}
\input{LaTeX/sections/3_method}
\input{LaTeX/sections/4_experiments}
\input{LaTeX/sections/5_conclusion} 

\section*{Acknowledgements}
This research was partly supported by the Sichuan Science and Technology Program (2025JDDQ0008, 2024NSFJQ0035), and the Talents by Sichuan provincial Party Committee Organization Department.


%
%
\bibliographystyle{splncs04}
\bibliography{main}
\end{document}

%% file: preamble.tex
\usepackage{caption}
\usepackage{wrapfig}
\renewcommand{\figurename}{Fig.}
\renewcommand{\tablename}{Tab.}
\usepackage{multirow}
\usepackage{graphicx}
\usepackage{mdframed}

\usepackage{overpic}
\usepackage{enumitem}
\usepackage{amsmath}

\usepackage{booktabs}
\usepackage{placeins}
\usepackage{amsfonts}

\usepackage{xcolor}
\usepackage{colortbl}

\definecolor{lightgray}{rgb}{0.9,0.9,0.9}
\definecolor{darkred}{rgb}{0.7,0.1,0.1}
\definecolor{darkgreen}{rgb}{0.1,0.5,0.1}
\definecolor{diffred}{rgb}{0.7,0.0,0.0}

\newcommand{\boldparagraph}[1]{\vspace{0.5em}\noindent{\bf #1.}}

\renewcommand{\paragraph}[1]{\boldparagraph{#1}}

\usepackage{arydshln}



%% file: LaTeX/sections/0_abstract.tex
\begin{abstract}
This paper focuses on a challenging setting of simultaneously modeling geometry and appearance of hand-object interaction scenes without any object priors.
We follow the trend of dynamic 3D Gaussian Splatting based methods, and address several significant challenges.
To model complex hand-object interaction with mutual occlusion and edge blur, we present interaction-aware hand-object Gaussians with newly introduced optimizable parameters aiming to adopt piecewise linear hypothesis for clearer structural representation.
Moreover, considering the complementarity and tightness of hand shape and object shape during interaction dynamics, we incorporate hand information into object deformation field, constructing interaction-aware dynamic fields to model flexible motions. 
To further address difficulties in the optimization process, we propose a progressive strategy that handles dynamic regions and static background step by step.
Correspondingly, explicit regularizations are designed to stabilize the hand-object representations for smooth motion transition, physical interaction reality, and coherent lighting.
Experiments show that our approach surpasses existing dynamic 3D-GS-based methods and achieves state-of-the-art performance in reconstructing dynamic hand-object interaction.
\vspace{-0.5cm}
\end{abstract}

%% file: LaTeX/sections/1_introduction.tex
\section{Introduction}
Accurate reconstruction of hand–object interaction (HOI)
is vital for VR and robotics~\cite{handa2020dexpilot}, requiring precise shape modeling and interaction capture.  
Despite the apparent simplicity of daily actions like grasping or drinking, they involve complex contact dynamics and severe occlusions that remain challenging to model. 
\begin{figure}[!ht]
    \centering
    \includegraphics[width=\linewidth]{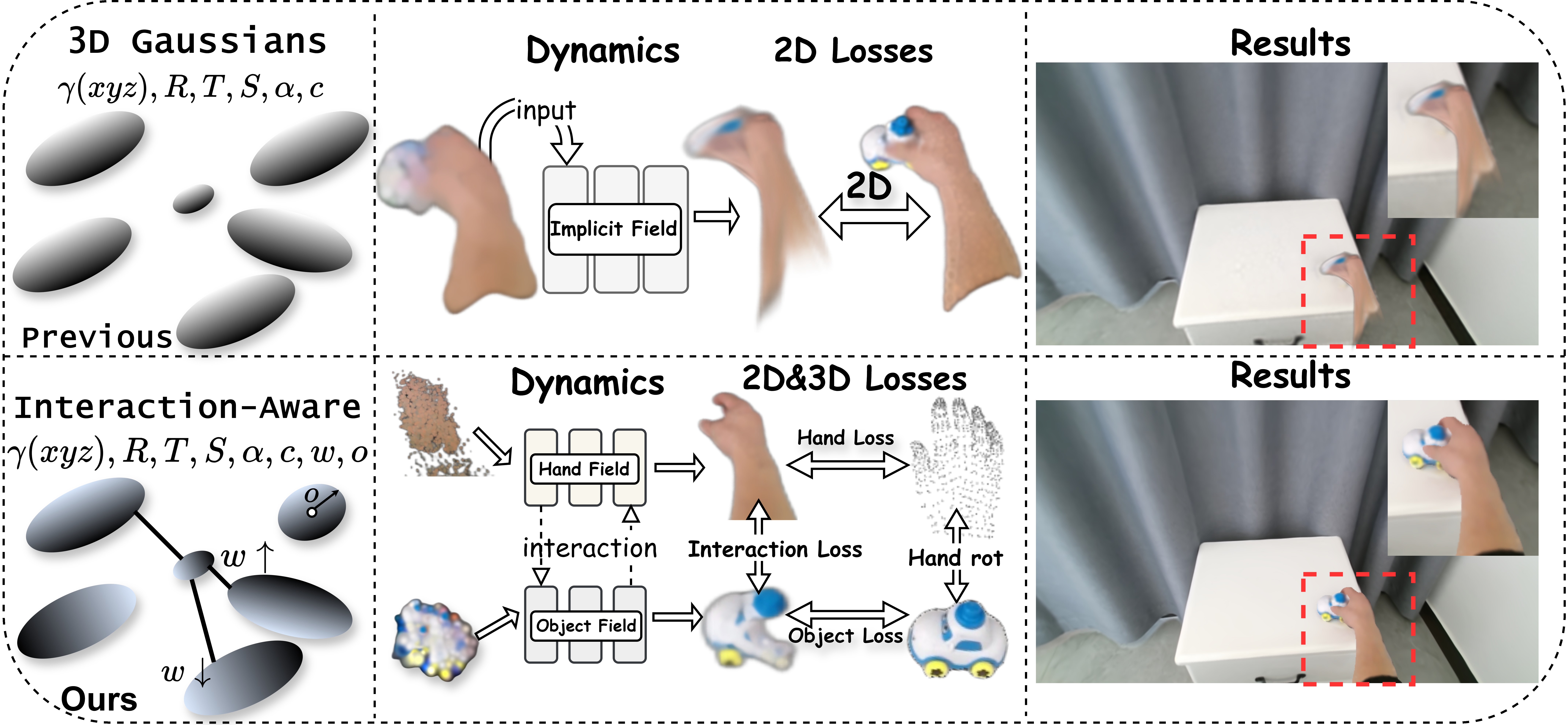}
    \caption{\textbf{Differences between traditional 3D Gaussian-based hand-object reconstruction and our interaction-aware modeling.}
    Conventional 3D Gaussian approaches model the entire HOI scene with a single, unified implicit field and rely primarily on 2D supervision. This design often leads to geometric ambiguities during close interactions—such as collapsed clearances, blurred contact boundaries, and non-physical merging of hand and object surfaces (top). In contrast, our method explicitly decouples hand and object representations into separate fields, introduces interaction-aware parameters ($\boldsymbol{w}, \boldsymbol{o}$) to modulate occlusion and edge sharpness, and leverages interaction-aware losses to preserve fine-grained spatial relationships, enabling accurate and disentangled dynamic reconstruction (bottom). 
   }
    \label{fig:fig1}
    \vspace{-0.6cm}
\end{figure}%
Previous works \cite{frankmocap_hand, grabnet} reconstruct HOI scenes by treating interactive objects as known, relying on specific object poses or templates. 
However, acquiring such poses or templates is costly, limiting their industrial applicability. 
Existing methods try to reduce the reliance on precise pose estimation and templates by several routes. 
\cite{qi2024hoisdf, bundlesdf} employs SDF-based approaches, integrating SDF to reconstruct dynamic hand-object scenes using neural networks, without the need for specific object priors. 
However, these methods focus primarily on geometry reconstruction without appearance.
With the advent of NeRF \cite{nerf}, some researchers \cite{liu2023hosnerf, zhang2023neuraldome} explore HOI scene reconstruction with both geometry and appearance by training implicit fields. 
However, due to the inefficiency of backward mapping-based ray-rendering algorithms \cite{nerf}, these methods require significant time and computational resources.
Recently, 3D Gaussian Splatting (3D-GS) \cite{3dgs} has demonstrated superior fidelity and speed in static scene reconstruction. 
Some works \cite{4dgs,scgs,deformable3dgs} attempt dynamic reconstruction using 3D-GS, but they struggle with complex HOI scenarios involving heavy occlusion and irregular rotations, failing to capture accurate interaction dynamics. 
Although EgoGaussian \cite{egogaussiandynamicsceneunderstanding} targets HOI reconstruction, it requires object pose estimation and only presents results for interactive objects without effective hand representation. 
In this work, we address the limitations of 3D-GS-based methods by proposing a model to simultaneously reconstruct the entire HOI scene without requiring any object priors.

To successfully handle such a practical setting presents significant challenges. 
First, drastic motions, mutual occlusion and blur during interaction cause misalignment and excessive overlap among Gaussians.
To address this, we model the interaction as a piecewise linear process and present a novel representation termed interaction-aware hand-object Gaussians.
It introduces two parameters over the traditional 3D-GS representation: weight $w$ and radius $o$. 
The weight $w$ balances motion smoothness and noise reduction, with smaller values indicating weak structural information or occlusion. 
The radius $o$ controls edge sharpness, where smaller values produce clearer contours. 
The combination of $w$ and $o$ effectively models the complex dynamic interaction, reducing blurring at interaction boundaries and enhancing visual quality. 
Second, previous methods \cite{scgs,deformable3dgs} use a single field to model Gaussian transformations, which is insufficient for capturing drastic and highly localized motions in HOI scenes, often leading to significant loss of fine-grained motion details. 
On the other hand, simply using separate fields for hand and object deformation overlooks their mutual interaction and geometric coupling during contact. 
To address this, we incorporate key-frame hand positions into the object field, enabling interaction-aware transformations that accurately capture dynamic changes caused by hand grasping.
Third, considering the flexible motions, irregular rotations, and frequent occlusions in HOI scenes, it is difficult to directly utilize traditional 3D-GS optimization \cite{3dgs,4dgs,deformable3dgs} to achieve decent rendering quality. 
To address this, we design explicit interaction-aware regularizations to explicitly stabilize the position and rotation of hand-object Gaussians.
Furthermore, we propose a progressive optimization mechanism to achieve physically realistic hand-object interaction, ensure smooth edge transitions, and enhance coherence in complex HOI scenes.

Our contributions are summarized as follows:
\begin{itemize}

    \item We propose a novel interaction-aware hand-object Gaussian representation to model HOI scenes without any object priors, effectively addressing mutual occlusion and edge blur during interactions.
    \item We enhance the object field with hand information to capture interaction-induced deformation. 
    \item We use a progressive optimization strategy with explicit 3D losses to benefit the fitting of the Gaussians during dynamic reconstruction.
    \item Experiments show that our approach surpasses state-of-the-art baselines, achieving superior performance in reconstructing dynamic HOI scenes. 
\end{itemize}

%% file: LaTeX/sections/2_related_work.tex
\section{Related Works}
\label{sec:rw}

\textbf{Hand Representation.}
Early approaches \cite{ganerated3dhand_moncular, realtime_handposeshpae_2hoi} focused on estimating 2D or 3D keypoints from images. The introduction of statistical hand models like MANO \cite{mano} revolutionized parametric hand representation by jointly encoding pose, shape, and 3D vertices. Recent method \cite{pushing_nerual_rendering_rgb_based_dense3dhandpose} typically employs regression networks to predict MANO parameters directly from images and optimize shape parameters for alignment. However, these methods suffer from error propagation: initial MANO inaccuracies accumulate downstream, causing cascading reconstruction errors. To mitigate this, we propose a hand field to decouple hand deformation from strict MANO parameter dependencies. Recent works have extended hand modeling using neural and Gaussian representations~\cite{manus, harp_Karunratanakul_2023_CVPR, DyTact_Cong_2025_ICCV}. In contrast, our work targets prior-free, monocular, egocentric HOI reconstruction without requiring multi-view inputs or personalized templates.

\textbf{Hand-Object Reconstruction.}\label{rw:rw2}
Reconstructing hand-object interaction from video remains a significant challenge in computer vision and graphics. Previous works fall into two categories. The first~\cite{honnotate, capturinghand, interhand2.6m, camera_space_hand_mesh} reconstructs hand-object interactions from multiview sources by fitting objects into 2D images using 3D object priors. However, these methods heavily rely on accurate 3D priors, which are costly to obtain. The second stream \cite{frankmocap_hand, arctic, learning_joint_hmo} pre-learns object templates to reduce reliance on priors. For example, the MANO model \cite{mano} represents the canonical hand space, with linear blending skinning driving the hand template. EgoGaussian~\cite{egogaussiandynamicsceneunderstanding} reconstructs egocentric interaction scenes using 3D Gaussian~\cite{3dgs} by separating dynamic objects from the static background.  
However, it is sensitive to object pose, probably failing under inaccurate poses, and excludes interacting hands from reconstruction. As a result, it misses the complete hand–object interaction context. Recent work BIGS~\cite{BIGS_On_2025_CVPR} reconstructs hand-object interactions from monocular video using 3D-GS and a diffusion prior, but assumes a known object mesh and excludes the background. In contrast, our method is category-agnostic, requires no object priors, and reconstructs the full hand-object interaction scene. A growing line of work leverages text-to-image diffusion priors to 
hallucinate unseen object geometry under heavy occlusion. Ye~\etal\ 
\cite{diffhoi_Ye_2023_ICCV}, G-HOP~\cite{G-HOP_Ye_2024_CVPR}, and 
MagicHOI~\cite{MagicHOI_Wang_2025_ICCV} share our prior-free goal but trade flexibility 
for diffusion-induced regularization, while our approach reconstructs 
HOI scenes without any external generative model. For articulated objects 
with parametric priors, BimArt~\cite{BimArt_Zhang_2025_CVPR} and 
SyncDiff~\cite{SyncDiff_He_2025_ICCV} target an orthogonal problem space. We 
position our work as complementary: prior-free reconstruction of egocentric 
HOI scenes involving rigid objects, without diffusion priors or object 
templates.

\textbf{Dynamic Scene Reconstruction.}\label{rw:rw3}
With the advent of NeRF \cite{nerf}, many works~\cite{dnerf, forwardflow_nvsds, hypernerf, nerf_spacetime_flowds, nerfds, nerfies, norigid_nerf} use MLPs to represent implicit spaces as deformation fields with temporal information. However, their extensive training time limits practical applicability. 
3D Gaussian Splatting (3D-GS) \cite{3dgs} emerges as a promising alternative for scene reconstruction. 
Methods like \cite{4dgs, deformable3dgs, scgs, kratimenos2024dynmf, luiten2024dynamic_3d_gaussian, katsumata2024compact_dynamic_gaussian, liu2025generalizableHOI, phygaussian, lin2024gaussian_flow, song2025coda_4dgs, lei2025mosca, zhang2025dynamic2dgs, stearns2024dynamic_marbles, xu2024grid4d} explore dynamic reconstruction using 3D-GS. 
For instance, \cite{deformable3dgs} uses MLPs to learn Gaussian position offsets per timestamp, which increases training time. 
\cite{scgs} introduces sparse control points to deform Gaussians, but in hand-object interaction (HOI) scenes, redundant points lead to inaccuracies and image tearing, failing to capture intricate interactions. 
These methods \cite{4dgs, deformable3dgs, scgs} input all Gaussians into a single MLP, which struggles to accurately model complex interactive motions. 
To overcome significant challenges posed by HOI scenes, our method introduces a novel interaction-aware hand-object Gaussian representation, with adaptive losses and a progressive optimization strategy.

%% file: LaTeX/sections/3_method.tex
\section{Method}
\label{sec:method}
Our goal is to reconstruct dynamic hand-object interaction (HOI) scenes from RGB egocentric videos at \textit{arbitrary timestamps} without relying on any object \textit{shape priors}.
We utilize three implicit fields to model the dynamic HOI scenes: the hand field $\mathcal{F}_\text{H}$ and object field $\mathcal{F}_\text{O}$ to approximate the shape of the dynamic HOI region, as well as the background field $\mathcal{F}_\text{BG}$ to create a clean background and facilitate subsequent joint optimization.
Separate modeling allows capturing clear hand-object appearance and stable background scene in drastically changing dynamic scenarios.
First, by treating hand and object modeling differently, significant occlusions could be solved via more detailed supervision. 
Second, backgrounds require low-frequency updates, while hand-object interactions demand high-frequency modeling. 
Meanwhile, collaborating with such dynamic implicit fields, we adaptively improve the Gaussian Splatting representation for HOI scenarios, addressing occlusion and contour clarity issues during the interaction. 
In optimization, we utilize only the hand's MANO parameters predicted by an off-the-shelf hand tracker~\cite{mano_hand_trajectye2025predicting}, which incurs negligible computational overhead compared to Gaussian optimization (typically $<3\%$ of total runtime). These pose estimates provide coarse 3D hand guidance to accelerate convergence without requiring any object priors. To ensure physically plausible interactions, we further introduce an interaction loss. A progressive and collaborative optimization framework is then devised to achieve high-quality HOI scene reconstruction through this lightweight 3D supervision and our interaction-aware representation.
\begin{figure*}[ht!]
    \centering
    \includegraphics[width=1\textwidth]{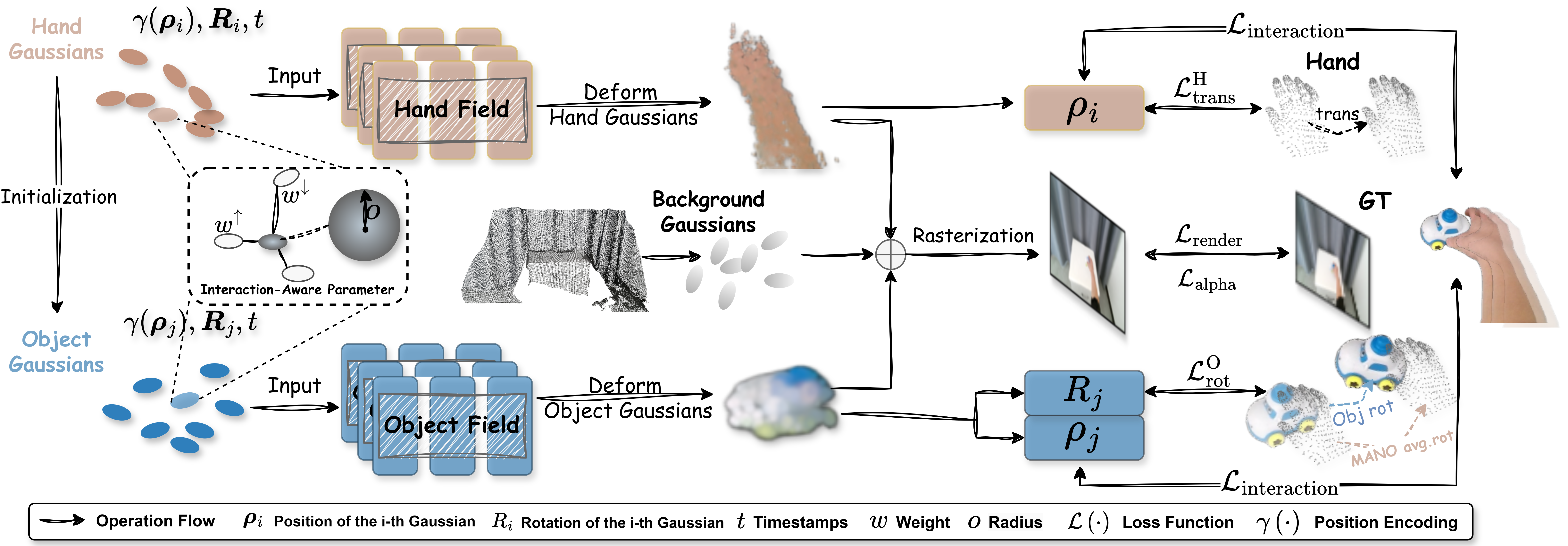}
    \caption{\textbf{Overview of interaction-aware hand-object Gaussians. }
    We propose a novel framework for reconstructing dynamic HOI scenes from RGB videos without object shape priors. 
    The framework consists of three components: (1) \textit{Specialized Implicit Fields}: separate hand, object, and background fields disentangle dynamic interactions, with hand/object fields capturing high-frequency deformations and occlusions (leveraging hand information for object's interaction-aware deformation) while the background field maintains low-frequency stability; 
    (2) \textit{Interaction-aware Gaussian}: enhances representation with adaptive weights $w$ and radius $o$ to address contour ambiguity and occlusions; 
    (3) \textit{Progressive Optimization}: combines explicit supervision with physical interaction constraints for efficient convergence.} 
    \label{fig:fig2}
    \vspace{-0.6cm}
\end{figure*}%

\subsection{Preliminaries: Deformable Gaussian Splatting}
\label{ssec:subhead1}
3D Gaussian Splatting (3D-GS) \cite{3dgs} represents 3D scene features using five parameters: position, transparency, spherical harmonic coefficients, rotation, and scaling.  
3D-GS explicitly defines each 3D Gaussian ellipsoid in space using a covariance matrix $\Sigma$ and the position vector \(\boldsymbol{\rho}\), as shown in the following equation:
\begin{equation}
\mathcal{G}(\boldsymbol{x})=\frac{1}{(2 \pi)^{\frac{3}{2}}|\Sigma|^{\frac{1}{2}}} \mathrm{exp}{(-\frac{1}{2}(\boldsymbol{x}-\boldsymbol{\rho})^{\top} \Sigma^{-1}(\boldsymbol{x}-\boldsymbol{\rho}))},
\end{equation}
here the $\Sigma$ matrix can be decomposed into a rotation $\boldsymbol{R}$ and a scaling $\boldsymbol{S}$ by \(\Sigma = \boldsymbol{RSS}^{\top} \boldsymbol{R}^{\top}\). 

Recent works \cite{deformable3dgs,4dgs,scgs} combine 3D-GS with deformation fields for dynamic scenes, using an MLP to warp points from canonical to target space:
\begin{equation}
\begin{aligned}
    \label{eq:eq4}
    \mathcal{F}_\theta(\boldsymbol{x}, \boldsymbol{t}) &= (\delta\boldsymbol{x}^t, \delta\boldsymbol{r}^t, \delta\boldsymbol{s}^t), 
    \\
    \boldsymbol{x}^t &= \boldsymbol{x}+\delta\boldsymbol{x}^t.
\end{aligned}
\end{equation}
Each initialized 3D Gaussian's center $\boldsymbol{x}$ is input to deformation field $\mathcal{F}_\theta$, which outputs time-dependent offsets ($\delta\boldsymbol{x}^t$, $\delta\boldsymbol{r}^t$, $\delta\boldsymbol{s}^t$) to wrap canonical Gaussians to target space. 

\subsection{Interaction-Aware Hand-Object Gaussians}
\label{ssec:subhead2}
To effectively capture the complex spatiotemporal motion in the Hand-Object Interaction (HOI) scene, we propose to decompose the dynamic HOI scene into three sets of Gaussians and model each part individually.
Moreover, we improve traditional 4D Gaussian representations to overcome the issues representing complex HOI motions in dynamic scenarios.
Traditional 4D Gaussians \cite{4dgs, deformable3dgs,scgs} tend to neglect mutual influences between different interacted Gaussians.
Moreover, it is hard to depict contour edges in interaction process, leading to texture drift and edge blur. 
Inspired by \cite{scgs}, we introduce two additional learnable parameters weight $\boldsymbol{w}\in \mathbb{R}^{+}$ and radius $\boldsymbol{o}\in \mathbb{R}^{+}$, forming a novel representation termed the Interaction-Aware Hand-Object Gaussian $\mathcal{G}_\text{HO}$. 
$\mathcal{G}_\text{HO}$ focuses on interaction-aware modeling via: 
(1) Weight $\boldsymbol{w}$ smooths the motion and reduces noise, it models mutual occlusion during interactions, where a small weight $\boldsymbol{w}$ indicates weak structural information and occlusion; 
(2) Radius $\boldsymbol{o}$ captures edge details, where a small radius $\boldsymbol{o}$ corresponds to sharper geometric contours near edges; 
(3) Since $\boldsymbol{w}$ is larger near the current Gaussian and smaller farther from the edge, the combination of weight $\boldsymbol{w}$ and radius $\boldsymbol{o}$ effectively handles edge blurring between hand-object interactions and the background. $\mathcal{G}_\text{HO}$ is given as follows: 
\begin{equation}
    \mathcal{G}_\text{HO} = \left\{\boldsymbol{x}_i\boldsymbol{y}_i\boldsymbol{z}_i,\boldsymbol{R}_{i},\boldsymbol{T}_i, \boldsymbol{S}_i, \boldsymbol{\alpha}_i,\boldsymbol{c}_i, \boldsymbol{w}_i, \boldsymbol{o}_i\right\}. 
\end{equation}
Due to different characteristics of hand motions, object motions and background scenes, we introduce each component’s dynamics separately below.

\textbf{Hand Gaussians.}~Hand Gaussians~$\mathcal{G}_\text{H}$ has the same optimizable parameters with $\mathcal{G}_\text{HO}$, and is modeled with a hand-implicit field $\mathcal{F}_\text{H}$ to capture the time-varying transformation of hand motion.
This field $\mathcal{F}_\text{H}$ takes the timestamp $t$ and the canonical position $\left(\boldsymbol{x}_i \boldsymbol{y}_i \boldsymbol{z}_i\right)$ of the $i$-th hand Gaussian as inputs. 
Since our setting requires modeling dynamic HOI scenes at any time from any view pose, we use $t$ as the input and construct the following formula (details provided in the supplementary material):
\begin{equation}
\Delta\mathcal{G}_\text{H} = \mathcal{F}_\text{H}\left\{\gamma\left(\boldsymbol{x}_i\boldsymbol{y}_i\boldsymbol{z}_i\right), \gamma\left(t\right)\right\} ,  
\label{eq:hand_field}
\end{equation}
$\gamma(\cdot)$ is positional encoding~\cite{posencoding1, deformable3dgs}. Adding $ \text{noise}_\text{smooth} $~\cite{deformable3dgs} to $\gamma(t)$ prevents oversmoothing and retains hand details while fitting coarse geometry.

\textbf{Object Gaussians.}~Hand-object interactions often cause deformations or occlusions (e.g., holding).  
To enhance the ability of the object field $\mathcal{F}_\text{O}$ to capture interaction-aware deformations, we introduce hand position as an additional input to the object field.   
This overcomes the limitation of implicit methods~\cite{4dgs, deformable3dgs}, which generate global offsets without explicit hand-object interaction modeling.
The object field $\mathcal{F}_\text{O}$ takes both hand and object positions as inputs, formulated as follows (details in supplementary material):  
\begin{equation}
     \Delta \mathcal{G}_\text{O} = \mathcal{F}_\text{O} \left\{\gamma\left(\left(\boldsymbol{x}^k_i\boldsymbol{y}^k_i\boldsymbol{z}^k_i\right) \oplus \left(\boldsymbol{x}^k_j\boldsymbol{y}^k_j\boldsymbol{z}^k_j\right)\right), \gamma\left(t\right)\right\}.
\end{equation}
Here, $\oplus$ concatenates hand $(\boldsymbol{x}^k_i, \boldsymbol{y}^k_i, \boldsymbol{z}^k_i)$ and object $(\boldsymbol{x}^k_j, \boldsymbol{y}^k_j, \boldsymbol{z}^k_j)$ positions with the canonical Gaussian position at key-frame $k$—the moment just before hand-object interaction.  
The object-implicit field $\mathcal{F}_\text{O}$ predicts time-varying offsets $\Delta\mathcal{G}_\text{O}$ at timestamp $t$, and uses linear annealing of $\text{noise}_\text{smooth}$~\cite{deformable3dgs} to stabilize training.

\textbf{Background Gaussians.}
We construct background Gaussians to better capture the smoothness and static nature of the background and avoid the unstable dynamic changes of the background Gaussian distribution caused by the interaction of foreground hand-object, which will affect the rendering quality of the background \cite{scgs, 4dgs}. 
Background Gaussians $\mathcal{G}_\text{BG}$ are based on the Deform3DGS model \cite{deformable3dgs}. 
Their positions change over time, as formulated in Eq.~\eqref{eq:eq4}, using the background-implicit field $\mathcal{F}_\text{BG}$ with timestamps $t$.

\subsection{Explicit Interaction-Aware Regularizations}
\label{ssec:subhead3}
2D regularization is to constrain pixel errors in image space. However, this is insufficient due to significant occlusion and drastic motion in hand-object interaction scenes. 
To enable Gaussians to accurately and efficiently model complex hand-object interactions, besides 2D supervision, we introduce explicit 3D regularizations from interaction priors.
Critically, these losses require no object shape or pose priors.
We only leverage a lightweight, off-the-shelf hand tracker~\cite{mano_hand_trajectye2025predicting} to provide coarse 3D hand guidance. 
These include object, hand, and interaction losses to stabilize the rotation and transformation of interaction-aware hand-object Gaussians, ensuring physically plausible dynamics without relying on any object-specific assumptions. 

\textbf{Hand Loss.}
Hand movement in HOI scenes is fast, making dynamic Gaussian fitting much slower and more challenging. 
Since MANO vertices explicitly represent the position of each point, we design a hand loss to optimize the translation of hand Gaussians. 
To track their translation, we use a single Chamfer Distance (CHD) to supervise Gaussian translation in 3D space, we compute its distance to the nearest vertex on the MANO $\mathcal{V}_h$.
This loss measures the distance from each hand Gaussian to its closest point on the MANO vertices, encouraging the Gaussians to populate the hand surface, formulated as follows:
\begin{equation}
\label{eq:hand_loss}
\mathcal{L}^\text{H}_\text{trans} = \frac{1}{N} \sum_{i=1}^N \min_{\mathbf{v} \in \mathcal{V}_h}\left\|\left(\boldsymbol{x}_i\boldsymbol{y}_i\boldsymbol{z}_i\right)-\left(\boldsymbol{x}_v\boldsymbol{y}_v\boldsymbol{z}_v\right)\right\|_2^2,
\end{equation}
where $\left(\boldsymbol{x}_i\boldsymbol{y}_i\boldsymbol{z}_i\right)$ denotes the $i$-th Gaussian position, $\left(\boldsymbol{x}_v\boldsymbol{y}_v\boldsymbol{z}_v\right)$ represents the filtered points within MANO vertices' range (addressing arm-hand discrepancies).

\textbf{Object Loss.}
Since object motion—both translation and rotation—tends to follow hand motion, we regularize the object field using hand cues. While translation is implicitly aligned via spatial proximity, rotation often suffers from non-physical flipping, especially in passive contacts. We thus introduce a hand-guided rotation loss $\mathcal{L}^{\text{O}}_{\text{rot}}$ that aligns object Gaussians with the hand’s dominant rotational trend.

In tightly coupled interactions (e.g., grasping), object rotation typically follows the hand’s dominant rotational trend.  
We compute a global prior $\mathbf{R}_{\text{hand}}^{\text{target}}(t) \in \mathrm{SO}(3)$ by averaging relevant MANO joint rotations via SVD-based averaging~\cite{2013Rotation_average}.  
To apply regularization only during contact, we modulate the loss with the interaction-aware weight $\boldsymbol{w}_j^\text{O}$ of each object Gaussian $j$, which is small under occlusion/non-contact and large when engaged. The contact-aware weight is:
\begin{equation}
    \omega_j(t) = \sigma(\boldsymbol{w}_j^\text{O}),
\end{equation}
where $\sigma$ is the sigmoid function. The final loss is:
\begin{equation}
    \mathcal{L}^{\text{O}}_{\text{rot}} = \mathbb{E}_{t} \left[ 
    \frac{1}{N} \sum_{j=1}^{N} 
    \omega_j(t) \left\| \log\left( (\mathbf{R}_{\text{hand}}^{\text{target}}(t))^{-1} \mathbf{R}_{t,j}^{\text{obj}} \right) \right\|^2 
\right],
\end{equation} $\mathbf{R}_{t,j}^{\text{obj}} \in \mathrm{SO}(3)$ is the predicted rotation of the $j$-th object Gaussian at timestamp $t$, and $\log(\cdot)$ denotes the logarithmic map from $\mathrm{SO}(3)$ to its Lie algebra $\mathfrak{so}(3)$. This metric-aware penalty suppresses implausible rotations while preserving local deformation freedom.

\textbf{Interaction Loss.}
Reconstruction of grasping interactions often suffers from edge blurring and mutual occlusion of Gaussians.
To regularize the physical reality, we introduce the self-supervised Chamfer distance between hand and object Gaussians. 
Our approach models the hand and object separately, explicitly defining their positions. 
This allows us to introduce an interaction loss to ensure proper grasping, formulated as follows:
\begin{equation}
\label{eq:eq11}
\begin{split}
\mathcal{L}_\text{interaction} = 
&\frac{1}{\max(|C_H|, \epsilon)} \sum_{i \in C_H} \min_{j \in C_O} \| \mathbf{p}_i - \mathbf{p}_j \|_2^2 \\
&+ \frac{1}{\max(|C_O|, \epsilon)} \sum_{j \in C_O} \min_{i \in C_H} \| \mathbf{p}_i - \mathbf{p}_j \|_2^2,
\end{split}
\end{equation}
where $\epsilon = 10^{-6}$ avoids division by zero when no contacts are detected.  
While this loss promotes hand–object proximity, it does not prevent interpenetration.  
We therefore use a separate penetration loss (supplementary material) that penalizes overlapping or overly close Gaussians from the hand and object. This loss ensures physical realism and visual quality by minimizing hand–object Gaussian distances while avoiding overlap.

\subsection{Progressive Optimization}
\label{ssec:subhead4}
In the Hand-Object Interaction (HOI) scene, complex rotations, translations, and occlusions are common. 
Directly optimizing all Gaussians leads to slow convergence and positional misalignment. 
To address these issues, we propose a progressive optimization strategy for learning individual implicit fields, which operates in five phases as below: initialization, warm-up, HOI refinement, background optimization, and collaborative reconstruction.

\textbf{Initialization.}
The MANO vertices~\cite{mano} provide a coarse initialization for hand geometry, obtained from an off-the-shelf hand tracker~\cite{mano_hand_trajectye2025predicting} and used solely to bootstrap $\mathcal{G}_\text{H}$. In contrast, for the object, we do not assume any shape, category, or 3D bounding box prior. Instead, $\mathcal{G}_\text{O}$ is randomly initialized by uniformly sampling 3D points within an expanded axis-aligned bounding box (AABB) of the MANO vertices. For the background, $\mathcal{G}_\text{BG}$ is initialized from SfM-based sparse point clouds.

\textbf{Warm-up.}
During the warm-up phase, we use the proposed 3D losses besides the fundamental 2D losses.
For the hand field, we employ $\mathcal{L}^\text{H}_{\text{trans}}$ to guide the deformation of hand Gaussians $\mathcal{G}_\text{H}$, ensuring alignment with the target pose.
For the object field, to stabilize interaction-aware transformations, we use $\mathcal{L}^\text{O}_\text{rot}$.
During the warm-up phase, we periodically apply gradient-based density adjustments \cite{3dgs} to optimize the initial Gaussian distribution.

\textbf{HOI Refinement.}
We adaptively refine Gaussians by assigning each Gaussian $i$ a learnable weight $\boldsymbol{w}_i$ and radius $\boldsymbol{o}_i$, $\boldsymbol{o}_i$ controls its local influence range.  
The final refinement weight for the $k$-th nearest neighbor of Gaussian $i$ is obtained by: (1) computing spatial proximity weights $\boldsymbol{w}^{\text{spatial}}_{ik}$ for the $K$ nearest neighbors via a RBF kernel on distance $\boldsymbol{d}_{ik}$ and $\boldsymbol{o}_i$ (Eq.~\ref{eq:spatial_weight}), (2) normalizing these weights to sum to one (Eq.~\ref{eq:normalized_spatial_weight}), and (3) modulating them with a global importance weight $\sigma(\boldsymbol{w}_i)$ (Eq.~\ref{eq:final_refinement_weight}).  
This allows joint learning of global importance $\boldsymbol{w}_i$ and local context.

\begin{equation}
\label{eq:spatial_weight}
\boldsymbol{w}^{\text{spatial}}_{ik} = \exp\left(-\frac{\boldsymbol{d}_{ik}^2}{2 \boldsymbol{o}_i^2}\right), \quad k \in \mathcal{N}_K(i),
\end{equation}
where $\mathcal{N}_K(i)$ denotes the set of $K$ nearest neighbor Gaussians for the $i$-th Gaussian, $\boldsymbol{d}_{ik}$ is the Euclidean distance between the centers of Gaussians $i$ and $k$, and $\boldsymbol{o}_i$ is the learnable radius parameter associated with Gaussian $i$.
\begin{equation}
\label{eq:normalized_spatial_weight}
\hat{\boldsymbol{w}}^{\text{spatial}}_{ik} = \frac{\boldsymbol{w}^{\text{spatial}}_{ik}}{\sum_{j \in \mathcal{N}_K(i)} \boldsymbol{w}^{\text{spatial}}_{ij}}.
\end{equation}
The refined weight of Gaussian $i$’s $k$-th neighbor is:
\begin{equation}
\label{eq:final_refinement_weight}
\hat{\boldsymbol{w}}^k_i = \sigma(\boldsymbol{w}_i) \cdot \hat{\boldsymbol{w}}^{\text{spatial}}_{ik}, \quad \text{for} \quad k \in \mathcal{N}_K(i),
\end{equation}
where $\sigma(\cdot)$ is the sigmoid function ensuring $\boldsymbol{w}_i \in (0, 1)$. 

Additionally, we query the hand implicit field $\mathcal{F}_{\text{H}}$ and the object implicit field $\mathcal{F}_{\text{O}}$ to obtain their respective rotation matrices $\left(\Delta \boldsymbol{R}_{6D} \in \mathbb{R}^{6}\right) \rightarrow \left(\Delta \boldsymbol{R} \in \mathbb{R}^{3 \times 3}\right)$ (see supplementary material) and translation offset $\Delta(\boldsymbol{x}^t_k\boldsymbol{y}^t_k\boldsymbol{z}^t_k)$. Using a linear blend of local rigid transformations inspired by LBS~\cite{LBSsumner2007embedded}, we refine the pose of the hand-object Gaussians as follows:
\begin{equation}
\begin{aligned}
\label{eq:eq13}
    \boldsymbol{T}^t_k & = (\boldsymbol{x}^t_k\boldsymbol{y}^t_k\boldsymbol{z}^t_k) + \Delta (\boldsymbol{x}^t_k\boldsymbol{y}^t_k\boldsymbol{z}^t_k), \\
    \boldsymbol{\rho}^t_i & = \sum_{k \in \mathcal{N}_K(i)}\hat{\boldsymbol{w}}^k_{i}\left(\Delta \boldsymbol{R}^t_k((\boldsymbol{x}_i\boldsymbol{y}_i\boldsymbol{z}_i)-(\boldsymbol{x}^t_k\boldsymbol{y}^t_k\boldsymbol{z}^t_k)) + \boldsymbol{T}^t_k\right).
\end{aligned}
\end{equation}
Here, $(\boldsymbol{x}_i\boldsymbol{y}_i\boldsymbol{z}_i)$ denotes the position of the $i$-th Gaussian in canonical space, and $\boldsymbol{\rho}^t_i$ represents the deformed position of the $i$-th Gaussian at timestamp $t$.

\textbf{Background Optimization.}
We pretrain $\mathcal{G}_\text{BG}$ for a fixed number of iterations, performing periodic density adjustments~\cite{3dgs} to ensure a clean background initialization.

\textbf{Collaborative Reconstruction.}
In the final stage, $\mathcal{F}_\text{H}$, $\mathcal{F}_\text{O}$, and $\mathcal{F}_\text{BG}$ independently deform their Gaussians into a shared target space, enabling full HOI scene reconstruction at any timestamp $t$. 
Both hand and object Gaussians employ HOI refinement (Eq.~\eqref{eq:eq13}) to update their parameters. 
The optimization is supervised by interaction constraints $\mathcal{L}_\text{interaction}$ (Eq.~\eqref{eq:eq11}) and 2D regularization terms. 
This stage ensures physically plausible occlusion relationships, smooth edge transitions, and lighting coherence, improving both the geometric fidelity of reconstructed shapes and the temporal smoothness of their motion dynamics.

%% file: LaTeX/sections/4_experiments.tex
\section{Experiments}
To validate our approach, we conduct comprehensive comparisons with state-of-the-art baselines \cite{4dgs,deformable3dgs,scgs} for dynamic scene reconstruction on both HOI4D~\cite{hoi4d} and HO3D~\cite{honnotate} datasets. Additionally, we compare with HOLD~\cite{fan2024hold} and BIGS~\cite{BIGS_On_2025_CVPR}, two specialized methods for hand-object interaction reconstruction, on the HO3D dataset. Following~\cite{egogaussiandynamicsceneunderstanding}, we evaluate pure translation and translation-rotation using alternate-frame testing to assess extrapolation to novel interactions. Metrics include PSNR, SSIM~\cite{psnr&ssim}, and LPIPS~\cite{lpips}. We further perform full-frame evaluation for completeness (Table~\ref{tab:quantitative_hoi4d} and~\ref{tab:quantitative_ho3d}, Ours*). All experiments run on a single NVIDIA RTX 3090 (24G), achieving optimal performance in 21,000 iterations (1h20m training time).

\textbf{Implementation Details.} We employ $K$ nearest neighbors for refinement and deformation, with the key-frame $k$ set to the timestamp just before hand–object contact.  
Both Gaussians and the deformation model are optimized using Adam.  
Hyperparameters, schedules, and auxiliary losses (penetration, momentum, 2D) are provided in the supplementary material. HOI4D~\cite{hoi4d} provides RGB-D videos with frame-level hand–object poses and masks; we evaluate on two purely translational and two translation–rotation scenes.  
HO3D~\cite{honnotate} offers real-world 3D pose annotations for actions like pickup and rotation.  
We use camera~4 from HO3D and select four translation–rotation sequences for egocentric reconstruction.  

\input{LaTeX/tables/hoi4d_results_new}
\input{LaTeX/tables/ho3d_results_new}
\begin{figure*}[!htb]
    \centering
    \includegraphics[width=\textwidth]{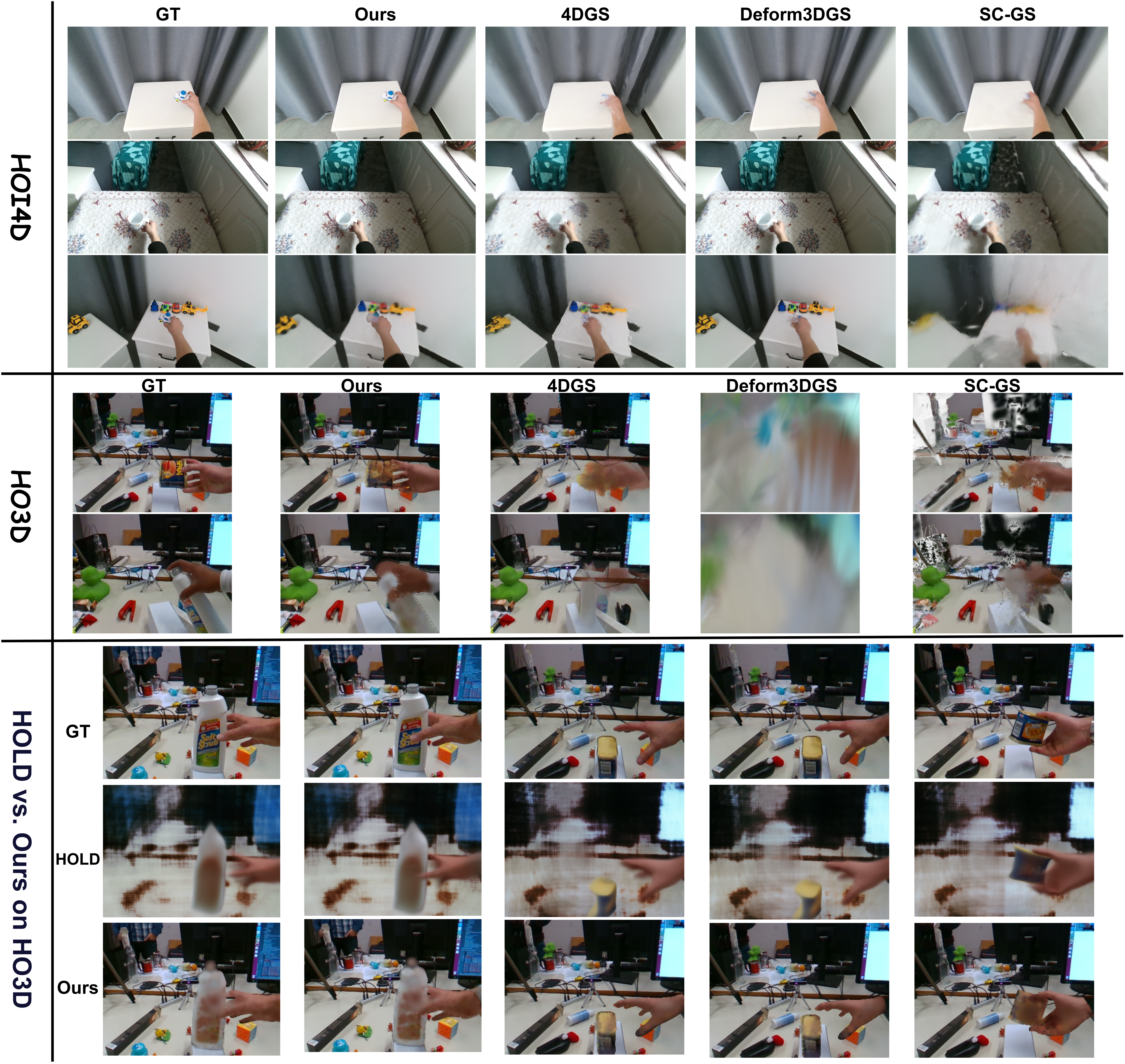}
    \caption{\textbf{Qualitative comparison of our approach and the baseline methods.}~We present reconstructions from our model and SOTA baselines (4DGS \cite{4dgs}, Deform3DGS \cite{deformable3dgs}, SC-GS \cite{scgs}) on HOI4D and HO3D datasets.} 
    \label{fig:fig3}
    \vspace{-0.5cm}
\end{figure*}%

\begin{figure}[!ht]
    \centering
    \includegraphics[width=0.95\textwidth]{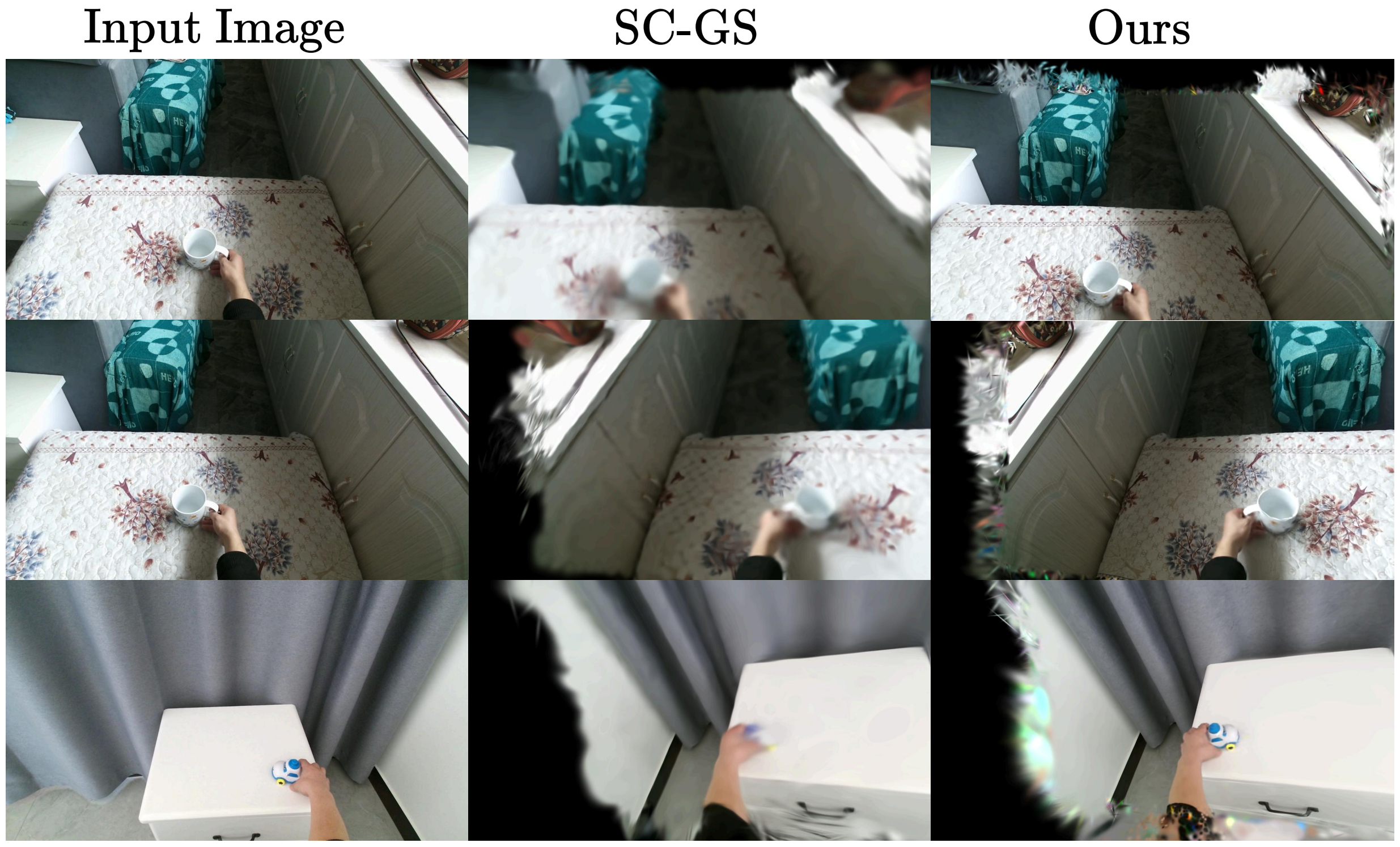}
    \caption{Extrapolated novel view synthesis of our approach and SC-GS~\cite{scgs}. 
    Our method shows cleaner renderings from novel viewpoints with considerable change, whereas SC-GS suffers from noticeable artifacts.} 
    \label{fig:fig_nvs}
\end{figure}%

\subsection{Quantitative Comparisons}

\textbf{HOI4D Dataset.}~We compare against 4DGS~\cite{4dgs}, Deform3DGS~\cite{deformable3dgs}, and SC-GS~\cite{scgs} using official code and original HOI4D resolution (Table~\ref{tab:quantitative_hoi4d}). 
4DGS is sensitive to initialization and underperforms in HOI settings.  
Deform3DGS and SC-GS, relying on a single deformation field, fail under occlusion and fast motion. 
Our interaction-aware, progressively optimized model overcomes these issues, reducing occlusion artifacts and blur while preserving hand-object geometry.  
We achieve a +9\% PSNR gain in translation scenes and improve rotation-heavy scene PSNR from 23.57\,dB (Deform3DGS) to 24.16\,dB. 
\\
\textbf{HO3D Datasets.}~We downsample all input frames to half resolution for efficient processing of large-scale sequences.  
Although HO3D~\cite{honnotate} provides camera parameters and hand-object pose estimates, inherent inaccuracies and pose errors adversely affect the performance of all methods (Table~\ref{tab:quantitative_ho3d}).  
4DGS is highly sensitive to input noise; SC-GS’s sparse control points fail to model background–foreground interactions; and Deform3DGS suffers most due to HO3D’s pose errors (Appendix~D of~\cite{deformable3dgs}), causing non-convergence.  
HOLD~\cite{fan2024hold}, designed for geometry rather than view synthesis, lags behind 3DGS-based methods. BIGS~\cite{BIGS_On_2025_CVPR} reports poor metrics because it reconstructs only foreground hand-object without the background (foreground-only results: $\dagger$ in Table~\ref{tab:quantitative_ho3d}). Our approach outperforms all baselines in full-scene reconstruction.

\subsection{Qualitative Comparisons}
As shown in Fig.~\ref{fig:fig3}, our approach surpasses 4DGS \cite{4dgs}, Deform3DGS \cite{deformable3dgs}, SC-GS \cite{scgs} and HOLD~\cite{fan2024hold} in both appearance and shape. In the HOI4D scene, baselines fail to handle Gaussian offsets under dynamic lighting and interaction, while our interaction-aware representation preserves shadows and shapes. Isolated background Gaussians improve contrast and dark details. For the HOI4D scene, separate hand-object modeling and 3D losses effectively constrain interaction-aware deformations, with collaborative reconstruction smoothing motion and occlusion.  
In the HO3D scene featuring irregular flipping, rotation, and finger flexibility—4DGS falters under noisy or inaccurate data and complex motion.  
SC-GS loses fine details in interaction zones due to sparse control points and handles occlusions poorly.  
Deformable3DGS, sensitive to pose errors (Appendix~D of~\cite{deformable3dgs}), fails to converge on HO3D as errors amplify.  
HOLD reconstructs hand and object geometry but produces low-quality full-scene renderings.  
Our method uses $w$ and $o$ to reduce occlusion and blur, achieving superior rendering quality on HO3D and HOI4D.

\subsection{Ablation Study}
\input{LaTeX/tables/ablation_study_new_sty}
\begin{figure}[!htb]
    \centering
    \vspace{-0.3cm}
    \includegraphics[width=0.95\textwidth]{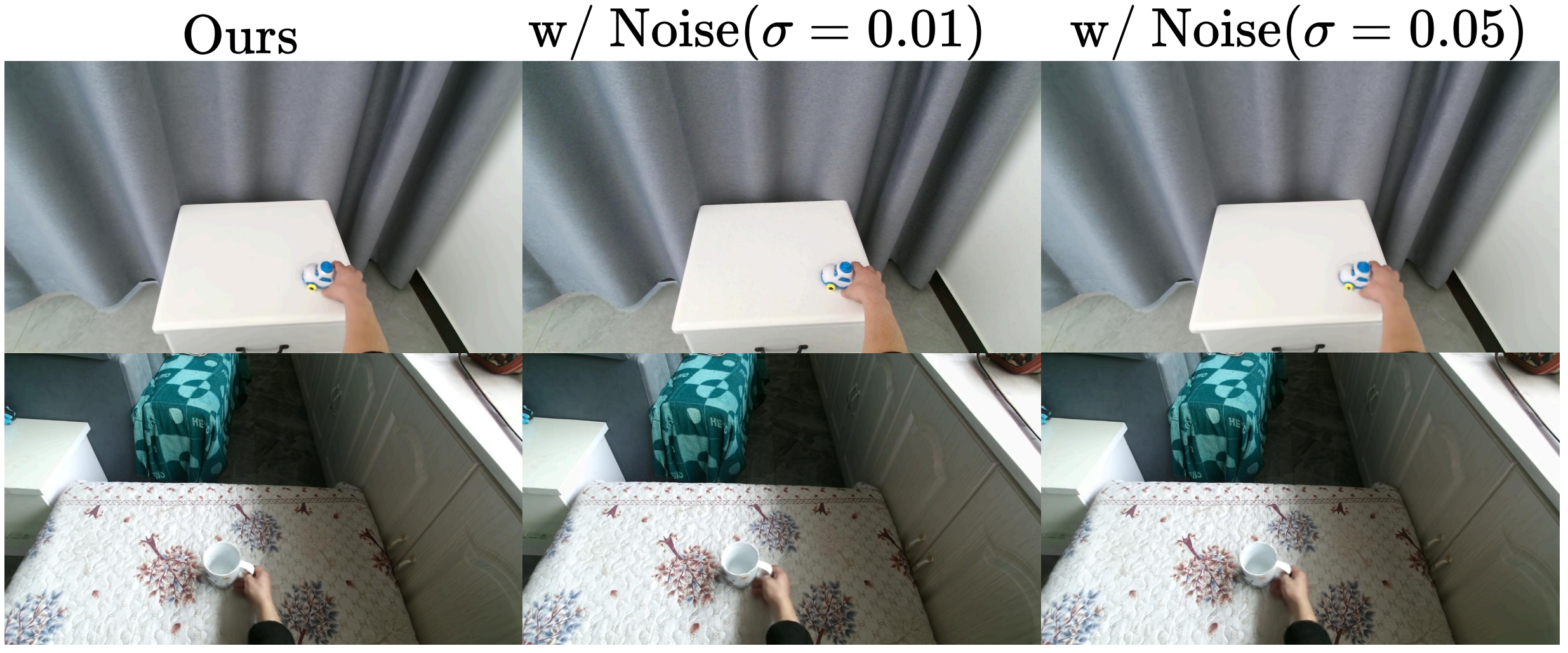}
    \caption{Our method maintains consistently high rendering quality across different noise levels, showing strong robustness to initialization errors.} 
    \label{fig:noise}
    \vspace{-0.3cm}
\end{figure}%
\begin{figure}[!htb]
    \centering
    \vspace{-0.1cm}
    \includegraphics[width=0.95\textwidth]{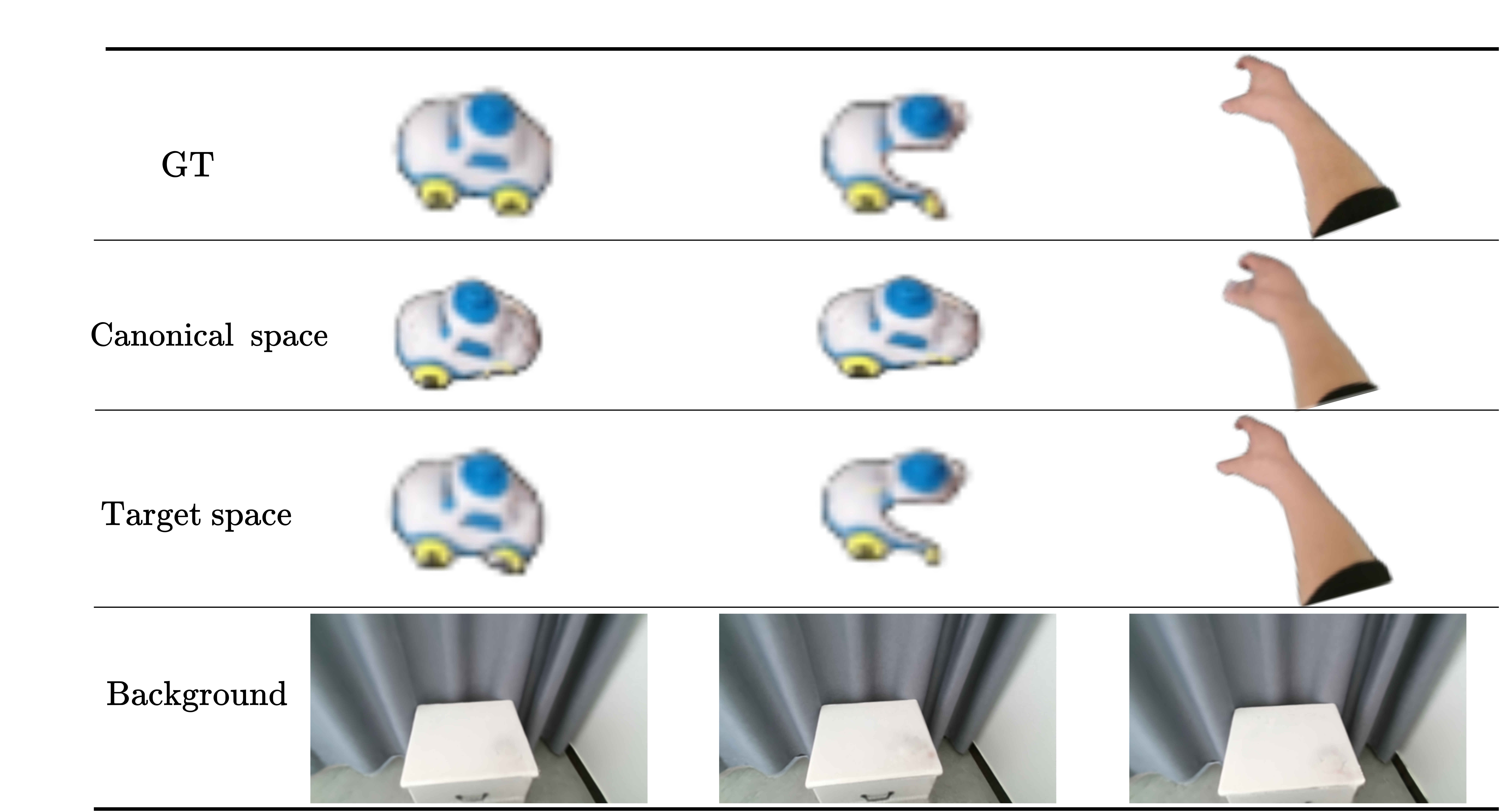}
    \caption{Disentangled rendering of hand, object, and background. 
Our method reconstructs each component with high fidelity while maintaining coherent interaction.} 
    \label{fig:decompose_render}
    \vspace{-0.6cm}
\end{figure}%
\noindent
Table~\ref{tab:ablation_new_style} reports ablation studies on HOI4D-Scene~1, evaluating the removal of HOI refinement, and the interaction-aware losses and module.  
We also evaluate robustness to imperfect initialization by adding Gaussian noise $\mathcal{N}(0, \sigma^2)$ to the randomly sampled initial object positions with $\sigma = 0.01$ and $0.05$. Table~\ref{tab:ablation_new_style} and Fig.~\ref{fig:noise}, our full model achieves near noise-free rendering quality under different noise levels. Removing HOI refinement degrades PSNR/SSIM/LPIPS by $2.2\%$/$1.1\%$/$11.4\%$; ablating object, hand, or interaction losses causes drops of ($4.6\%$, $1.1\%$, $8.6\%$), ($1.5\%$, —, $5.7\%$), and ($3.5\%$, $1.1\%$, $14.3\%$), respectively. To validate our interaction-aware design, we conducted an ablation study by eliminating both the field parameters and their associated training scheme from our framework. Tab.~\ref{tab:ablation_new_style} clearly shows the performance drop when the interaction-aware module is absent.

%% file: LaTeX/tables/hoi4d_results_new.tex
\begin{table*}[ht]
\centering
\setlength\tabcolsep{3pt}
\definecolor{diffred}{rgb}{0.7,0.0,0.0}
\begin{tabular}{@{}lccccccc@{}}
\toprule
\textbf{Methods} &
  \multicolumn{3}{c}{\textbf{Translation}} &
  \multicolumn{1}{c}{} &
  \multicolumn{3}{c}{\textbf{Translation\&Rotation}} \\
\cmidrule(lr){2-4} \cmidrule(lr){6-8}
&
  PSNR $\uparrow$ & SSIM $\uparrow$ & LPIPS $\downarrow$ &&
  PSNR $\uparrow$ & SSIM $\uparrow$ & LPIPS $\downarrow$ \\
\midrule

4DGS~\cite{4dgs}  
& 24.86\textcolor{diffred}{\scriptsize -5.46} & 0.80\textcolor{diffred}{\scriptsize -0.13} & 0.47\textcolor{diffred}{\scriptsize +0.18} 
& & 
\textit{23.68}\textcolor{diffred}{\scriptsize -0.48} & 0.85\textcolor{diffred}{\scriptsize -0.01} & 0.39\textcolor{diffred}{\scriptsize +0.02} \\

Deform3DGS~\cite{deformable3dgs}  
& \textit{26.33}\textcolor{diffred}{\scriptsize -3.99} & \textit{0.87}\textcolor{diffred}{\scriptsize -0.06} & \textbf{0.29}\textcolor{diffred}{\scriptsize \phantom{+}0.00} 
& & 
23.57\textcolor{diffred}{\scriptsize -0.59} & \textbf{0.89}\textcolor{diffred}{\scriptsize +0.03} & \textbf{0.28}\textcolor{diffred}{\scriptsize -0.09} \\

SC-GS~\cite{scgs}  
& 25.08\textcolor{diffred}{\scriptsize -5.24} & 0.84\textcolor{diffred}{\scriptsize -0.09} & \textit{0.46}\textcolor{diffred}{\scriptsize +0.17} 
& & 
17.32\textcolor{diffred}{\scriptsize -6.84} & 0.71\textcolor{diffred}{\scriptsize -0.15} & 0.48\textcolor{diffred}{\scriptsize +0.11} \\

Ours 
& \textbf{30.32} & \textbf{0.93} & \textbf{0.29} 
& & 
\textbf{24.16} & \textit{0.86} & \textit{0.37} \\

\midrule

Ours* 
& 33.03\textcolor{diffred}{\scriptsize +2.71} & 0.95\textcolor{diffred}{\scriptsize +0.02} & 0.27\textcolor{diffred}{\scriptsize -0.02} 
& & 
24.02\textcolor{diffred}{\scriptsize -0.14} & 0.85\textcolor{diffred}{\scriptsize -0.01} & 0.39\textcolor{diffred}{\scriptsize +0.02} \\

\bottomrule
\end{tabular}%
\caption{\textbf{Quantitative comparison on HOI4D}~\cite{hoi4d}. Best and second-best results are \textbf{bolded} and \textit{italicized}, respectively. Differences (red, small font) are relative to the ``Ours'' row. Ours* denotes full-frame evaluation.}
\label{tab:quantitative_hoi4d}
\vspace{-1.2cm}
\end{table*}%

%% file: LaTeX/tables/ho3d_results_new.tex
\begin{table}[ht] 
\centering
\setlength\tabcolsep{8pt} 
\definecolor{diffred}{rgb}{0.7,0.0,0.0}
\begin{tabular}{@{}lccc@{}}
\toprule
\textbf{Methods} &
  \multicolumn{3}{c}{\textbf{Translation\&Rotation}} \\
\cmidrule(lr){2-4}
&
  PSNR $\uparrow$ & SSIM $\uparrow$ & LPIPS $\downarrow$ \\
\midrule

BIGS~\cite{BIGS_On_2025_CVPR}  
& 3.85\textcolor{diffred}{\scriptsize -21.34} 
& 0.24\textcolor{diffred}{\scriptsize -0.65} 
& 0.70\textcolor{diffred}{\scriptsize +0.55} \\

HOLD~\cite{fan2024hold}  
& 18.03\textcolor{diffred}{\scriptsize -7.16} 
& \textit{0.84}\textcolor{diffred}{\scriptsize -0.05} 
& 0.26\textcolor{diffred}{\scriptsize +0.11} \\

4DGS~\cite{4dgs}  
& 19.44\textcolor{diffred}{\scriptsize -5.75} & 0.82\textcolor{diffred}{\scriptsize -0.07} & \textit{0.25}\textcolor{diffred}{\scriptsize +0.10} \\

Deform3DGS~\cite{deformable3dgs}  
& 9.68\textcolor{diffred}{\scriptsize -15.51} & 0.36\textcolor{diffred}{\scriptsize -0.53} & 0.65\textcolor{diffred}{\scriptsize +0.50} \\

SC-GS~\cite{scgs}  
& \textit{20.37}\textcolor{diffred}{\scriptsize -4.82} & 0.80\textcolor{diffred}{\scriptsize -0.09} & 0.26\textcolor{diffred}{\scriptsize +0.11} \\

Ours 
& \textbf{25.19} & \textbf{0.89} & \textbf{0.15} \\

Ours* 
& 25.17\textcolor{diffred}{\scriptsize -0.02} & 0.89\textcolor{diffred}{\scriptsize \phantom{+}0.00} & 0.16\textcolor{diffred}{\scriptsize +0.01} \\
\midrule

BIGS$^\dagger$
& 24.51\textcolor{diffred}{\scriptsize -0.68} 
& 0.92\textcolor{diffred}{\scriptsize +0.03} 
& 0.07\textcolor{diffred}{\scriptsize +0.08} \\
Ours$^\dagger$ 
& 28.16\textcolor{diffred}{\scriptsize +2.97} & 0.95\textcolor{diffred}{\scriptsize +0.06} & 0.07\textcolor{diffred}{\scriptsize -0.08} \\
\bottomrule
\end{tabular}
\caption{\textbf{Quantitative comparison on HO3D}~\cite{honnotate}. Best and second-best results are \textbf{bolded} and \textit{italicized}. Differences (red font) are relative to ``Ours''. Ours* denotes full-frame evaluation. $\dagger$ denotes evaluation on hand and object regions only.}
\label{tab:quantitative_ho3d}
\vspace{-0.8cm}
\end{table}%








%% file: LaTeX/tables/ablation_study_new_sty.tex
\begin{table}[ht]
    \centering
    \small
    \setlength{\tabcolsep}{8pt}
    \renewcommand{\arraystretch}{1.2}
    \definecolor{diffred}{rgb}{0.7,0.0,0.0}
    \begin{tabular}{@{}lccc@{}}
        \toprule
        Methods & PSNR$\uparrow$ & SSIM$\uparrow$ & LPIPS$\downarrow$ \\
        \midrule
        w/o Interaction-Aware Module &
        28.76\textcolor{diffred}{\scriptsize -4.20} &
        0.91\textcolor{diffred}{\scriptsize -0.04} &
        0.40\textcolor{diffred}{\scriptsize +0.05} \\

        w/o HOI Refinement &
        32.23\textcolor{diffred}{\scriptsize -0.73} &
        0.94\textcolor{diffred}{\scriptsize -0.01} &
        0.39\textcolor{diffred}{\scriptsize +0.04} \\

        w/o Object Loss &
        31.45\textcolor{diffred}{\scriptsize -1.51} &
        0.94\textcolor{diffred}{\scriptsize -0.01} &
        0.38\textcolor{diffred}{\scriptsize +0.03} \\

        w/o Hand Loss &
        32.45\textcolor{diffred}{\scriptsize -0.51} &
        0.95\textcolor{diffred}{\scriptsize \phantom{+}0.00} &
        0.37\textcolor{diffred}{\scriptsize +0.02} \\

        w/o Interaction Loss &
        31.79\textcolor{diffred}{\scriptsize -1.17} &
        0.94\textcolor{diffred}{\scriptsize -0.01} &
        0.40\textcolor{diffred}{\scriptsize +0.05} \\
        
        w/ noise $\sigma=0.01$ &
        32.80\textcolor{diffred}{\scriptsize -0.16} &
        0.95\textcolor{diffred}{\scriptsize \phantom{+}0.00} &
        0.35\textcolor{diffred}{\scriptsize \phantom{+}0.00} \\

        w/ noise $\sigma=0.05$ &
        32.72\textcolor{diffred}{\scriptsize -0.24} &
        0.95\textcolor{diffred}{\scriptsize \phantom{+}0.00} &
        0.35\textcolor{diffred}{\scriptsize \phantom{+}0.00} \\
        
        \rowcolor{lightgray}
        Full Model &
        \textbf{32.96} & \textbf{0.95} & \textbf{0.35} \\
        \bottomrule
    \end{tabular}
    \caption{Ablation studies on HOI4D. All metrics are reported relative to the Full Model (differences in red, small font). Noise robustness results correspond to Fig.~\ref{fig:noise}.}
    \label{tab:ablation_new_style}
\end{table}%

%% file: LaTeX/sections/5_conclusion.tex
\section{Conclusion}
In this paper, we propose interaction-aware hand-object Gaussians with novel optimizable parameters, adopting piecewise linear hypothesis for a clearer structural representation. 
This approach effectively captures complex hand-object interactions, including mutual occlusion and edge blur. 
Leveraging the complementarity and tight coupling of hand and object shapes, we integrate hand information into the object deformation field, constructing interaction-aware dynamic fields for flexible motion modeling. 
To improve optimization, we propose a progressive strategy that separately handles dynamic regions and static backgrounds.
Additionally, explicit interaction-aware regularizations enhance motion smoothness, physical plausibility, and lighting coherence. 
Experiments show that our approach outperforms the baseline methods, achieving state-of-the-art results in reconstructing dynamic hand-object interactions.

\textbf{Limitations.}~As designed for interaction modeling, the workflow consists of progressive optimization stages, which could be unified upon the emergence of new stronger optimizer. Our method struggles with extreme cases (exceedingly rapid motion/complex trajectories, see Supplement), potentially addressable by integrating more interaction priors.